\PassOptionsToPackage{prologue,dvipsnames}{xcolor}
\documentclass[letterpaper, 10 pt, conference]{ieeeconf}  

\IEEEoverridecommandlockouts                              

\usepackage[T1]{fontenc}
\usepackage{graphics}
\usepackage{graphicx}
\usepackage{booktabs}
\usepackage{diagbox}
\usepackage{tikz}
\usepackage{lipsum} 

\usepackage{pifont}
\usepackage{amsmath}
\usepackage{amssymb}
\usepackage{threeparttable}
\usepackage[dvipsnames]{xcolor}
\usepackage{multirow}
\usepackage{hyperref}
\usepackage{cite}
\usepackage{bbding,pifont}

\newcommand{\moraidataset}{MORDA}
\newcommand{\shiftdataset}{SHIFT}
\newcommand{\opvdataset}{OPV2V}
\newcommand{\synthiadataset}{SYNTHIA-AL}

\newcommand{\virtualkittiseconddataset}{VKITTI2}
\newcommand{\viperdataset}{VIPER}
\newcommand{\waymo}{Waymo}

\newcommand{\kittidataset}{KITTI}
\newcommand{\nuscenesdataset}{nuScenes}
\newcommand{\aihubdataset}{AI-Hub}
\newcommand{\fasterrcnn}{Faster-RCNN}

\newcommand{\pointpillars}{PointPillars}
\newcommand{\centerpoint}{CenterPoint}
\newcommand{\ssn}{SSN}

\newcommand{\map}{mAP}
\newcommand{\nds}{NDS}

\newcommand{\reftable}[1]{Tab. \ref{#1}}
\newcommand{\reffigure}[1]{Fig. \ref{#1}}
\newcommand{\refsection}[1]{Sec. \ref{#1}}

\definecolor{crimson}{rgb}{0.86, 0.08, 0.24}
\definecolor{darkblue}{rgb}{0.0, 0.0, 0.55}

\newcommand*{\red}[1]{\textcolor{crimson}{#1}} 
 
\newcommand*{\blue}[1]{\textcolor{darkblue}{#1}}

\newcommand*{\improve}[1]{\textbf{\textcolor{LimeGreen}{+#1}}}
\newcommand*{\degrade}[1]{\textcolor{BrickRed}{\textendash#1}}

\newcommand{\eg}{e.g.,}
\newcommand{\ie}{i.e.,}
\newcommand{\xmark}{\text{\ding{55}}}
\newcommand{\RN}[1]{%
  \textup{\uppercase\expandafter{\romannumeral#1}}%
}

\newcommand{\realsource}{\texorpdfstring{$D_{Real}^{Src}$}}
\newcommand{\realtarget}{\texorpdfstring{$D_{Real}^{Trg}$}}
\newcommand{\synaux}{\texorpdfstring{$D_{Syn}^{Src+Trg}$}}
\newcommand{\moraisim}{MORAI SIM}

\title{\LARGE \bf
\moraidataset{}: A Synthetic Dataset to Facilitate Adaptation of \\ Object Detectors to Unseen Real-target Domain \\While Preserving Performance on Real-source Domain
}

\author{Hojun Lim$^{*}$, Heecheol Yoo$^{*}$, Jinwoo Lee, Seungmin Jeon, Hyeongseok Jeon$^{\dag}$
\thanks{$\ast{}$ These two authors contributed equally}
\thanks{$\dag{}$ Corresponding author}
\thanks{All authors are associated with MORAI Inc., Republic of Korea. Email: {\tt\small\{hjlim, hcyoo, jwlee, smjeon, hsjeon\}@morai.ai}}%
\thanks{This research was supported by a grant(code RS-2023-00233952) from R\&D Program funded by Ministry of Land, Infrastructure and Transport of Korean government.}
}

\begin{document}
\maketitle
\thispagestyle{empty}
\pagestyle{empty}

\begin{abstract}
Deep neural network (DNN) based perception models are indispensable in the development of autonomous vehicles (AVs). 
However, their reliance on large-scale, high-quality data is broadly recognized as a burdensome necessity due to the substantial cost of data acquisition and labeling. 
Further, the issue is not a one-time concern as AVs might need a new dataset if they are to be deployed to another region (real-target domain) that the in-hand dataset within the real-source domain cannot incorporate.  
To mitigate this burden, we propose leveraging synthetic environments as an auxiliary domain where the characteristics of real domains are reproduced. This approach could enable indirect experience about the real-target domain in a time- and cost-effective manner. 
As a practical demonstration of our methodology, \nuscenesdataset{} and South Korea are employed to represent real-source and real-target domains, respectively. That means we construct digital twins for several regions of South Korea, and the data-acquisition framework of \nuscenesdataset{} is reproduced. Blending the aforementioned components within a simulator allows us to obtain a synthetic-fusion domain in which we forge our novel driving dataset, \textit{\moraidataset{}: Mixture Of Real-domain characteristics for synthetic-data-assisted Domain Adaptation}.  
To verify the value of synthetic features that \moraidataset{} provides in learning about driving environments of South Korea, 2D/3D detectors are trained solely on a combination of \nuscenesdataset{} and \moraidataset{}. 
Afterward, their performance is evaluated on the unforeseen real-world dataset (\aihubdataset{}\footnote{This research (paper) used datasets from \textbf{High-precision data collection vehicle daytime city road data}. 
All data information can be accessed through \textbf{AI-Hub} (\url{http://www.aihub.or.kr} ).}) collected in South Korea. 
Our experiments present that \moraidataset{} can significantly improve mean Average Precision (\map) on \aihubdataset{} dataset while that on \nuscenesdataset{} is retained or slightly enhanced. 
Details on \moraidataset{} can be accessed at \href{https://morda-e8d07e.gitlab.io/}{https://morda-e8d07e.gitlab.io}.

\end{abstract}
\section{INTRODUCTION}

Deep-learning-based techniques have obtained growing attention and have become a major trend in challenging perception tasks of autonomous vehicles (AVs). Such dominance occurs across types of sensor modality: camera \cite{tian2019fcos, liu2020smoke, detr3d}, LiDAR \cite{lang2019pointpillars, zhu2020ssn, yin2021center}, RADAR \cite{bang2024radardistillboostingradarbasedobject}, and even combination of those \cite{kim2023crn, liu2022bevfusion, yan2023cross}. Despite their thrives in AVs, one inherent challenge is the dependency on a large-scale driving dataset. In general, building a dataset for perception tasks involves (1) equipping a car with the desired sensor suite, (2) deploying the car to the regions of interest to collect sensor data, and lastly, (3) generating the ground truth (GT) labels for sensor data according to labeling rules. As widely known, this entire process is time-consuming and cost-heavy. 
\begin{tikzpicture}[remember picture,overlay]
\node[anchor=south,yshift=10pt] at (current page.south) {\parbox{\textwidth}{\centering
\footnotesize
© 2025 IEEE. Personal use of this material is permitted. Permission from IEEE must be obtained for all other uses, in any current or future media, including reprinting/republishing this material for advertising or promotional purposes, creating new collective works, for resale or redistribution to servers or lists, or reuse of any copyrighted component of this work in other works.
}};
\end{tikzpicture}%

This challenge gets even more severe when deploying AVs that have been developed for one domain (real-source domain, \realsource{}) to another (real-target domain, \realtarget{}). In this context, one preferable choice for safety-critical AVs would be rebuilding the training dataset with GT labels for the novel domain. However, this method is indeed pricey. 

To alleviate the labeling cost, research on unsupervised domain adaptation (UDA) has been actively conducted. UDA methods assume that only the \realsource{} is labeled, and they exploit the knowledge from the annotated \realsource{} for better adaptation (\ie{} performance) of DNNs to the unlabeled \realtarget{} without additional labeling costs \cite{yang2021st3d, yang2021st3ddenoisedselftrainingunsupervised}. 
Nevertheless, dispatching the sensor-equipped vehicle to \realtarget{} for data collection is inevitable even in this scenario. 
Further, the engineering cost to synchronize raw data from sensors operating often at different frequencies is non-negligible.

\begin{figure}[tb]
  \centering
    \includegraphics[width=0.5\textwidth]{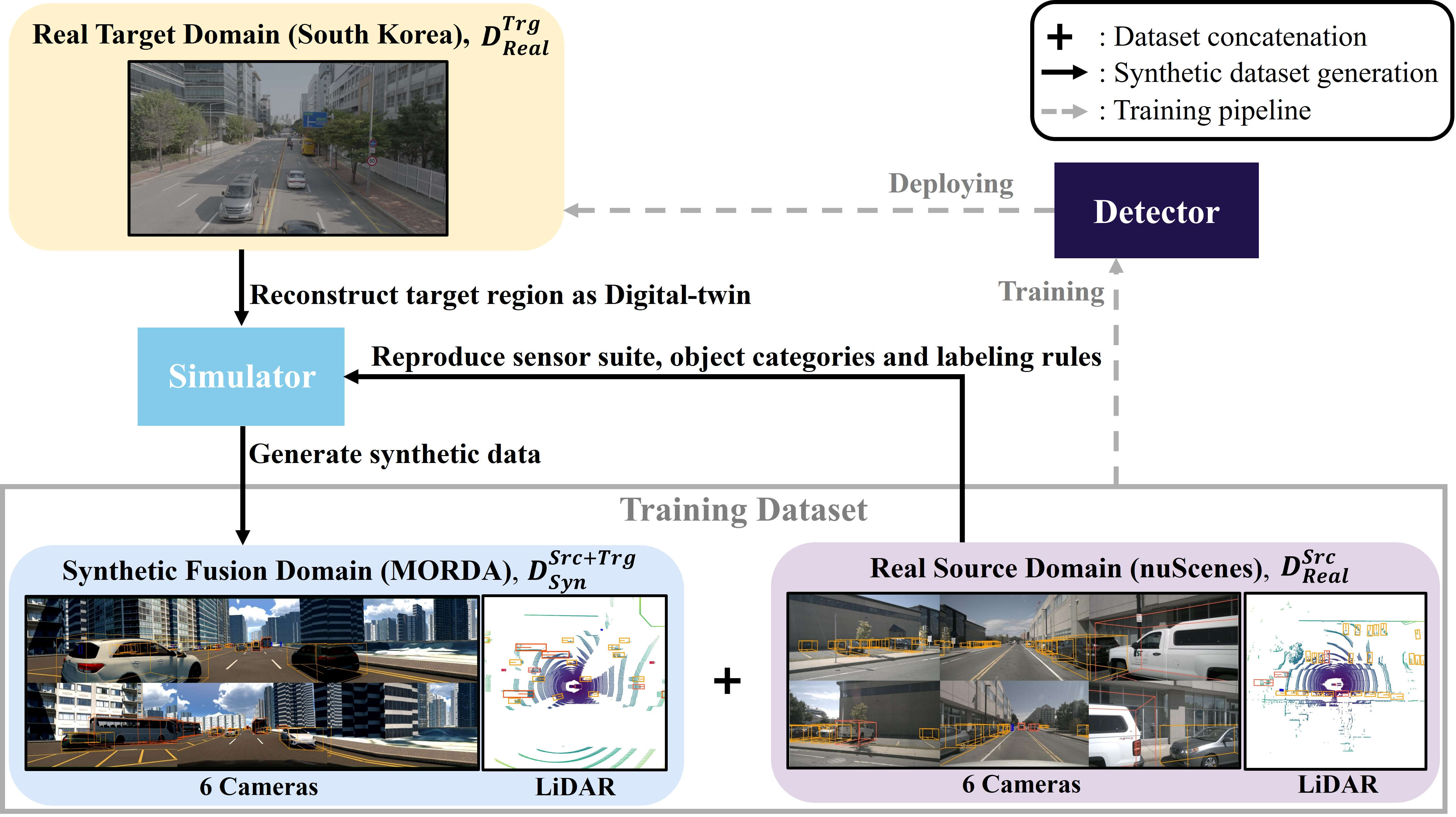}
  \vspace*{-6mm}
  \caption{Research overview: Cost-effective generation of virtual data (\synaux{}) which mimics real-world dataset that one could construct if the sensor configuration of \realsource{} were dispatched to the real-target domain \realtarget{}.
  The usefulness of our method is assessed by training object detectors on (\realsource{} + \synaux{}) and evaluating on unforeseen \realtarget{}.}
  \label{fig:intro}
  \vspace{-17pt}
\end{figure}

\begin{table*}[t]
    \caption{Comparison table for properties regarding sensor suite and BBox labels across synthetic datasets.}
    \vspace*{-2mm}
    \label{tab:synthetic_dataset_benchmark_table}
    \centering
    \begin{threeparttable}[b]
    \resizebox{0.8\textwidth}{!}{
    \begin{tabular}{l|cc|cc|c|ccc} 
    \hline
    \toprule[1.5pt]
    \multicolumn{1}{l|}{} & \multicolumn{5}{c|}{{\bfseries Sensor setup}} &  \multicolumn{3}{c}{{\bfseries BBox annotation}}\\ 
    \cline{2-9}
        & $\#$Camera\tnote{*} & Resolution & $\#$LiDAR & $\#$Channel & Sensor suite reference & DOF\tnote{\textdagger} & Road obstacle\tnote{\textdaggerdbl{}} & Frequency(Hz) \\
    \hline 
    \viperdataset{} \cite{Richter_2017}  & 1 & 1920$\times$1080 & \xmark & \xmark & - & 9 & \xmark &\textasciitilde{}15 \\
    \virtualkittiseconddataset{} \cite{vkitti2}& 5 & 1242$\times$375 & \xmark & \xmark & \kittidataset{} \cite{kitti} & 9 & \xmark & - \\ 
    \synthiadataset{} \cite{synthia_al}& 1 & 640$\times$480 & \xmark & \xmark & - & 7 & \xmark & 25 \\ 
    \shiftdataset{} \cite{shift2022} & 5 & 1200$\times$800 & 1 & 128 & - & 9 & \xmark & 10  \\
    \opvdataset \cite{OPV2V} & 4 & 800$\times$600 & 1 & 64 & - & 9 & \xmark & 10  \\
    \textbf{\moraidataset{} (Ours)} & 6 & 1600$\times$900 & 1 & 32 & \nuscenesdataset{} \cite{nuscenes} & 9 & \checkmark{} & 20 \\   
    \bottomrule[1.5pt]
    \end{tabular}}
    \begin{tablenotes}[para]
    \footnotesize
    \item [(*)] Only multi-view cameras are counted, and stereo cameras are not included. 
    \item [(\textdagger{})] Degree of freedom, 9 means 3D-BBox labels are represented with position (x, y, z), size (length, width, height), and orientation (roll, pitch, yaw). 7 denotes the orientation is represented only with yaw angle.
    \item [(\textdaggerdbl{})] Traffic cones and barriers. 
    \item [(-)] Relevant information is not available. 
    \end{tablenotes}
    \end{threeparttable}
    \vspace{-25pt}
\end{table*}

Recently, game engines and simulators have been increasingly employed to reduce the burden given that a large-scale dataset with accurate GT labels can be constructed in a time- and cost-efficient manner \cite{gta5_dataset, Richter_2017, shift2022, synthia_al, vkitti, vkitti2}. However, many synthetic driving datasets at present are not specifically designed to replicate particular real-world domains, including driving environments, sensor suites, object categories, and labeling policies. Consequently, inconsistencies between those characteristics of real-world and synthetic datasets might further widen the gap between them, potentially resulting in performance degradation of DNNs when trained on real and synthetic datasets combined.

Given the aforementioned constraints in UDA and existing synthetic datasets, the research objective of this paper is to develop a synthetic dataset that enhances the adaptation of DNNs to \realtarget{} while their performance on \realsource{} is not degraded. Note that this scenario is challenging as (1) \realtarget{} is completely unforeseen, meaning sensor data of the target domain is not used for training DNNs, and (2) the synthetic dataset should improve the generalization ability of DNNs as they need to perform well on both \realsource{} and \realtarget{}.

 We believe an adequate synthetic dataset for our purpose needs to be equipped with the following attributes; 
 (1) Virtual version of geographical features of driving environments in the target domain (\realtarget{}), which enables cost-effective indirect exposure of \realtarget{} to DNNs. 
 (2) Consistency with \realsource{} in terms of sensor suite and labeling process to keep the characteristics of created synthetic dataset paired with \realsource{}. 
 In this regard, we propose a methodology to generate a synthetic dataset in a synthetic-fusion domain (\synaux{}) where the mentioned aspects of \realsource{} and \realtarget{} are well integrated. 
 As a demonstration, we construct a novel synthetic dataset, \moraidataset{} which takes \nuscenesdataset{} for \realsource{} and South Korea for \realtarget{}. 
 Next, we evaluate the efficacy of our \moraidataset{} dataset in a challenging scenario, as illustrated in \reffigure{fig:intro}. Firstly, 2D/3D object detectors are trained solely on \nuscenesdataset{} (\realsource{}) combined with \moraidataset{} (\synaux{}). Afterward, they are evaluated on the unforeseen driving scenes of South Korea (\realtarget{}) present in \aihubdataset{} dataset.
 Our extensive experiments demonstrate that \moraidataset{} allows noteworthy detection-performance gain on \realtarget{} while successfully maintaining the performance on \realsource{} simultaneously. 
 In short, the main contributions are:
\begin{itemize}
    \item We present \moraidataset{}, a synthetic dataset featuring the virtual version of the real driving environments in South Korea and replicating key components of \nuscenesdataset{} in data acquisition and labeling.
    \item  Using \moraidataset{}, we show that utilizing \synaux{} could be a financially-viable method for detectors to adapt to \realtarget{} without performance degradation on \realsource{}.
\end{itemize}

\section{RELATED WORKS}
\subsection{Object Detection}
The field of object detection has shown surprising progress with DNNs in recent years, which aims to the accurate prediction of bounding box (BBox) to localize and classify objects within the sensor's field-of-view. The prosperity of DNN-based object detection applies to not only 2D but also challenging 3D detection tasks across various sensor modalities \cite{tian2019fcos, liu2020smoke, detr3d, lang2019pointpillars, zhu2020ssn, yin2021center, bang2024radardistillboostingradarbasedobject, kim2023crn, liu2022bevfusion, yan2023cross}. However, the inherent challenge of learning algorithms remains that the performance degradation occurs when DNNs are evaluated on a new domain which is unforeseen during training \cite{BenDavid2010ATO, shift2022}. Note that addressing such is imperative for AVs whose driving environments or conditions can often change, \eg{} deploying the developed AVs to another country (\realsource{}$\rightarrow$\realtarget{}).

\subsection{Unsupervised Domain Adaptation (UDA)}
To tackle the performance drop in new domains, numerous studies have concentrated on adapting DNNs trained on the source domain (w/ labels) to the target domain (w/o labels). 
Recent works in this field present notable advances even in challenging LiDAR-based 3D object detection. For example, \cite{yang2021st3d, yang2021st3ddenoisedselftrainingunsupervised, wozniak2024uada3dunsupervisedadversarialdomain} address the adaptation problem of \waymo{} \cite{Sun_2020_CVPR} $\rightarrow$ \nuscenesdataset{} \cite{nuscenes}, presenting notable improvement. 

Despite the striving of UDA, several difficulties remain to be resolved. 
Firstly, adversarial training schemes are often employed to extract domain-invariant features, expecting the learned representations to adapt seamlessly to the target domain \cite{vu2018advent, wozniak2024uada3dunsupervisedadversarialdomain}.
However, as noted in \cite{Zhao2019OnLI}, if there is a severe distribution mismatch between source and target domains, it can lead to insufficient generalization.
Secondly, sophisticated self-training schemes \cite{pan2020unsupervised, yang2020fdafourierdomainadaptation, yang2021st3d, yang2021st3ddenoisedselftrainingunsupervised}, which leverage knowledge obtained from the labeled source domain to generate pseudo labels for the target domain, naturally risk potential error propagation. 
Lastly, sensor data collection at the target regions is still required, which could be expensive in practice even without labeling. 

\begin{figure*}[t]
    \centering
    \includegraphics[width=\textwidth]{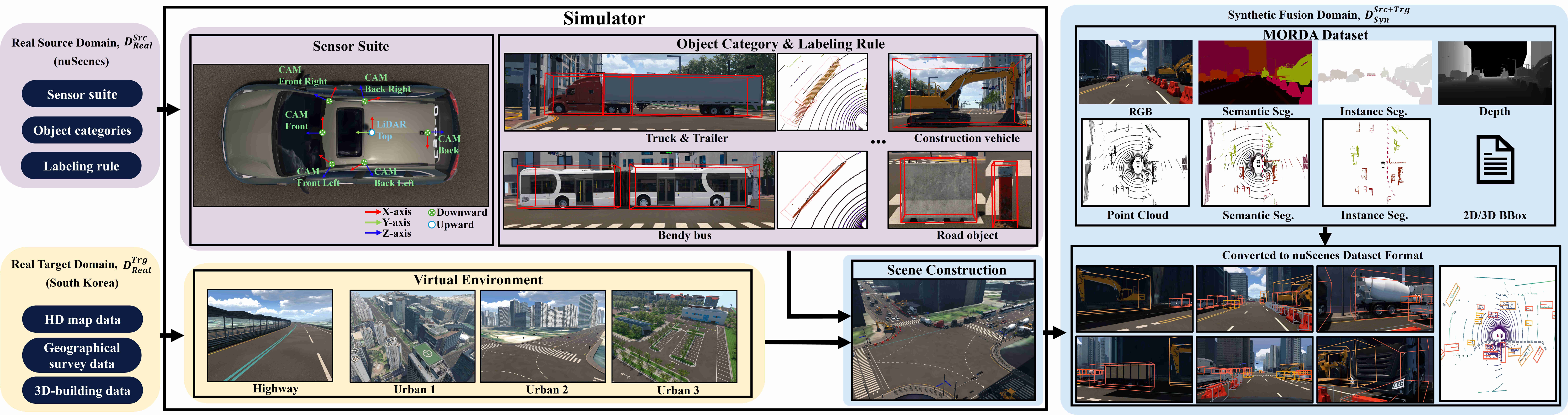}  
    \vspace*{-6mm}
    \caption{Architecture of the proposed method to generate \moraidataset{} from synthetic-fusion domain (\synaux{}).
     Digital-twin maps are leveraged to reflect locational features of \realtarget{}. Sensor suite and labeling rules from \realsource{} are implemented to suppress factors that could cause unexpected discrepancies in data distributions of \realsource{} and \synaux{}. Combining both, we ensure \synaux{} is well aligned with \realsource{} and \realtarget{}.
    }
    \label{fig:morai_dataset_generation_overview}
    \vspace{-15pt}
\end{figure*}

\subsection{Synthetic Driving Datasets}
Cost-efficient virtual environments have been actively employed to reduce the burden of constructing large-scale driving datasets with rich annotations.
For example, Unity\footnote{\url{http://unity3d.com/}} engine is used to generate Virtual KITTI 2  (\virtualkittiseconddataset{}) \cite{vkitti2}, and \synthiadataset{} \cite{synthia_al}. 
A modern game Grand Theft Auto \RN{5} is utilized to create GTA5 \cite{gta5_dataset} and VIPER \cite{Richter_2017}. 
Recently, open-source driving simulator CARLA \cite{pmlr-v78-dosovitskiy17a} has been employed to produce SHIFT \cite{shift2022} and \opvdataset{} \cite{OPV2V}.

Despite prevailing achievements of existing synthetic datasets, they are less suitable for scenarios where their role is to assist the adaptation of DNNs trained on \realsource{} to unobserved \realtarget{}. 
The reason is that many synthetic datasets are generated from their own sensor suite and virtual worlds with no real-world references. Therefore, there is no guarantee that the geological features of \realtarget{}, a target region for deploying AVs with DNNs, are present inside them. \virtualkittiseconddataset{} is an exception as it is a digital-version clone of a real-world dataset, \kittidataset{}. 
The mentioned characteristics of \virtualkittiseconddataset{} enable it to assess the usefulness of virtual worlds in terms of transferability to the real world. Nevertheless, this validation is limited to camera-related tasks, as LiDAR of \kittidataset{} is not implemented in \virtualkittiseconddataset{}. Furthermore, BBox labels for cyclists and pedestrians that exist in \kittidataset{} are absent, which makes it difficult to be used for learning about classes other than vehicles.

Our \moraidataset{} dataset benchmarks \nuscenesdataset{}, meaning it incorporates cameras, LiDAR, and 10 detection classes with accurate BBox labels for both modalities. \reftable{tab:synthetic_dataset_benchmark_table} compares \moraidataset{} with publicly available synthetic datasets.

\section{\moraidataset{} DATASET GENERATION}

\subsection{Overview}
The purpose of our method is to construct a dataset from a synthetic-fusion domain (\synaux{}) where DNNs (specifically, 2D/3D object detectors) can indirectly learn about the unforeseen \realtarget{} to enhance adaptability while retaining the performance on \realsource{} at which the model is trained. 
To accomplish the aim, we extract pivotal characteristics from \realsource{} and \realtarget{}.
Then, they are reproduced and blended within a simulator, creating our fusion domain. 
Lastly, our synthetic dataset (\moraidataset{}) is generated using the simulator. 
\reffigure{fig:morai_dataset_generation_overview} illustrates this procedure in detail.
As a showcase of the proposed methodology, we choose \nuscenesdataset{} and South Korea to represent \realsource{} and \realtarget{}, respectively. Next, the MORAI Simulator (\moraisim{}) \cite{morai} is employed to provide virtual environment where fundamental features like driving agents and GT-label generation are available.

\subsection{Reproduced Characteristics of Source Domain, \realsource{}}
We extract three components from \realsource{} and implement them in \moraisim{} to create \moraidataset{}: sensor suite, list of object categories for detectors to classify, and labeling rules to draw 3D BBoxes for each category. 
As \nuscenesdataset{} is employed for \realsource{}, its sensor suite (six 1600$\times$900 cameras and one 32-channel spinning LiDAR), object categories (10 detection classes), and annotation rules for 3D BBox are reproduced. 
The positions and orientations of individual virtual sensors adhere to the configuration of real ones in \nuscenesdataset{}. 
It is worth mentioning that 3D BBox labeling rules even for complicated categories that have two individual rigid sections (\eg{} bendy bus, truck-trailer) are implemented as visualized in \reffigure{fig:morai_dataset_generation_overview}. 
Further, the types of objects used for dataset generation include construction vehicles and road objects (barriers, traffic cones), which are rarely observed in other synthetic datasets at present.

\subsection{Reproduced Characteristics of Target Domain, \realtarget{}}
We want the geographical features of the real target domain, South Korea, to be reflected in the virtual environment so that our generated \moraidataset{} dataset could be enriched with them. 
To ensure this, we employ four digital-twin maps of South Korea available in \moraisim{}, which consist of one highway and three urban cities. 
Note that each digital-twin map was constructed upon HD map data, geographical survey data, and 3D building data collected from each counterparting region in South Korea so that the road surface, terrain, and surrounding static objects can be realistically replicated in the simulator.

\subsection{Data Creation in Synthetic Fusion Domain, \synaux{}}
By reproducing the mentioned real-world characteristics in the simulator, we construct \synaux{}, the synthetic-fusion domain where we hypothesize that the features from \realsource{} and \realtarget{} are appropriately blended. 
87 scenes are made in this domain, and for each scene, the ego vehicle drives on a predefined path at a certain place of a digital twin map to collect synthetic sensor data.  
The scenes are composed of two types: static and dynamic, accounting for 57 and 30.

The major difference between static and dynamic scenes is the presence of moving objects. 
In static scenes, objects other than the ego vehicle are stationary. 
However, their arrangement has been meticulously handcrafted to emulate parking lots, construction sites, and road environments filled with huge vehicles.
Hence, this type of scene has a larger number of trucks, trailers, construction vehicles, barriers, and traffic cones than the other type. 
In contrast to static scenes, the focus of dynamic scenes is to incorporate dense traffic flows with many moving objects into our dataset. 
To achieve this, we exploit the built-in traffic generator of \moraisim{}, which spawns various agents (\eg{} cars and motorcyclists) driving autonomously, to keep the road congested.\footnote{The traffic flow was set to reach E (operation near or at capacity) or F (breakdown in flow) of Level-of-service (LOS). Details regarding this can be found in \cite{manual2013ministry}.} 

While having the ego vehicle drive in a scene, the synchronized virtual sensors collect data at 20 Hz along with ground-truth (GT) labels. 
Iterating this process across all scenes yields our \moraidataset{} dataset enriched with the simulated geographical features of \realtarget{} and object-shape features of \realsource{}. 
Lastly and optionally, we convert \moraidataset{} to follow the dataset format of \nuscenesdataset{}. The lower-right corner of \reffigure{fig:morai_dataset_generation_overview} illustrates our converted dataset visualized using \nuscenesdataset{} development toolkit (nuscenes-devkit) \cite{nuscenes}. 
The reason for this conversion is detailed in \refsection{source_domain_od_data_preprocessing} 

\section{\moraidataset{} Dataset}

\subsection{Dataset Content}
\moraidataset{} comprises \textasciitilde{}37K frames where each frame consists of ego-vehicle pose data, six 1600$\times$900 images, and one 32-channel point cloud data with corresponding GT labels. 
The types of GT include pixel/point-level semantic and instance segmentation, 2D/3D BBox annotation, and pixel-wise depth values. 
All types of GT labels are generated independently by individual virtual sensors. 
Therefore, our dataset can be utilized for various vision tasks such as monocular/multi-view camera-based, lidar-based, fusion-based object detection, and segmentation tasks. 

\subsection{Comparison to \nuscenesdataset{} Dataset} \label{sec:comp_to_nusc}

\moraidataset{} delivers 1.6 million (M) 3D BBox labels for \textasciitilde{}37K frames, whereas \nuscenesdataset{} provides 40K frames with 1.4M BBoxes (only keyframes are counted excluding sweeps as they are not annotated). While both datasets present comparable scales of labeled frames, \moraidataset{} has \textasciitilde{}15$\%$ more BBox labels than \nuscenesdataset{}. Furthermore, our dataset presents denser temporal information of objects around ego vehicles as \moraidataset{}'s BBox labels are generated at 20 Hz while those of \nuscenesdataset{} are labeled at 2 Hz.

\begin{figure}[tpb]
  \centering
    \includegraphics[width=0.4\textwidth]{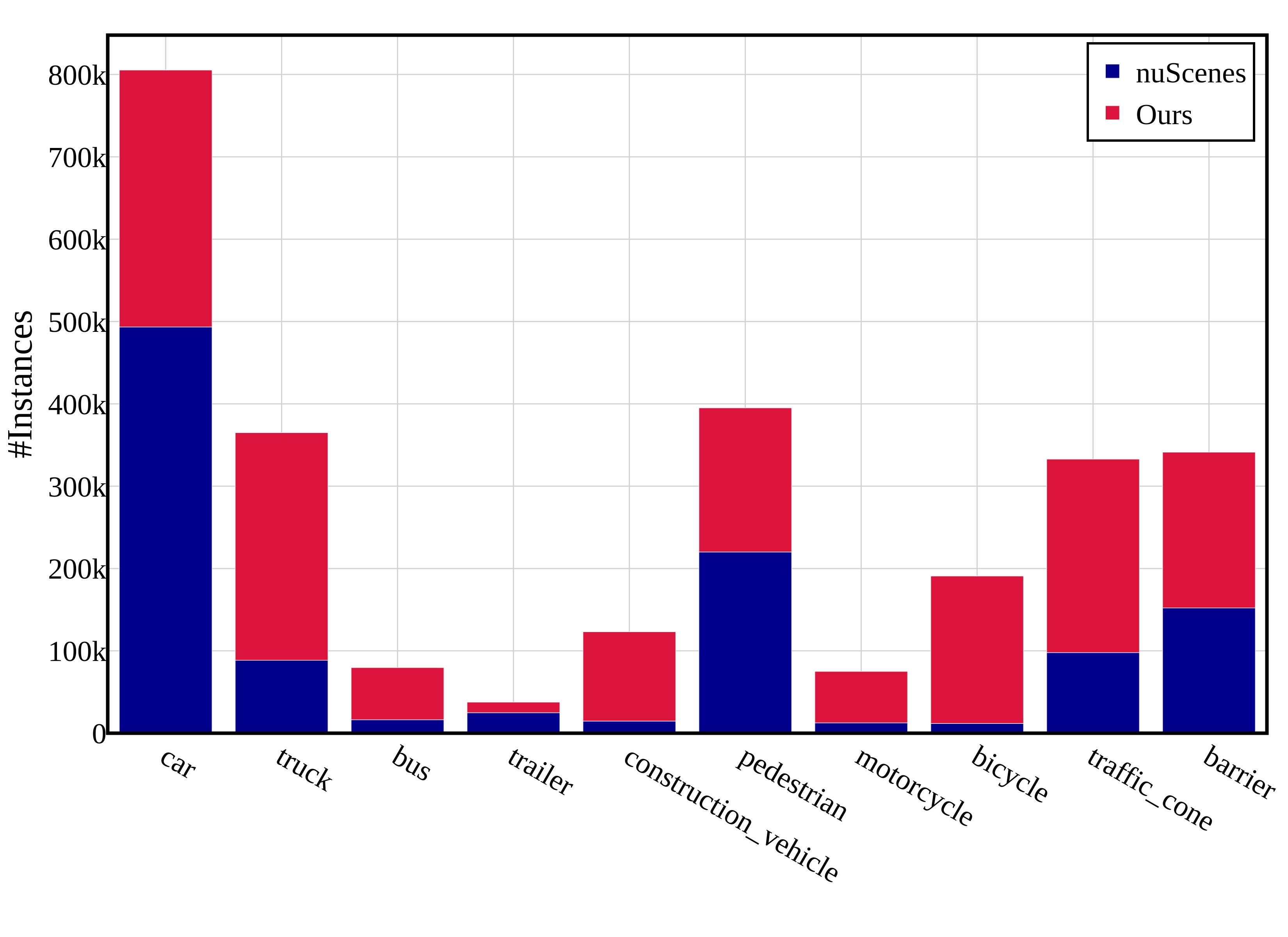}
  \vspace*{-5mm}
  \caption{Distribution of 3D-BBox annotations by category for \nuscenesdataset{} (\blue{blue}, train + val splits) and \moraidataset{} (\red{red}).} 
  \label{fig:morai_nusc_3dbox_distribution}
  \vspace{-5pt}
\end{figure}

\reffigure{fig:morai_nusc_3dbox_distribution} visualizes the count of 3D BBoxes for each class by dataset. 
As shown, their distributions largely differ, but \moraidataset{} supplements the shortage of 3D BBoxes for several classes in \nuscenesdataset{}. Specifically, the longtailness of rare classes (truck, bus, construction vehicle, trailer, motorcycle, and bicycles) in \nuscenesdataset{} is mitigated when 3D-BBox labels of respective classes in \moraidataset{} are concatenated. 

\begin{figure}[t]
  \centering
    \includegraphics[width=0.5\textwidth]{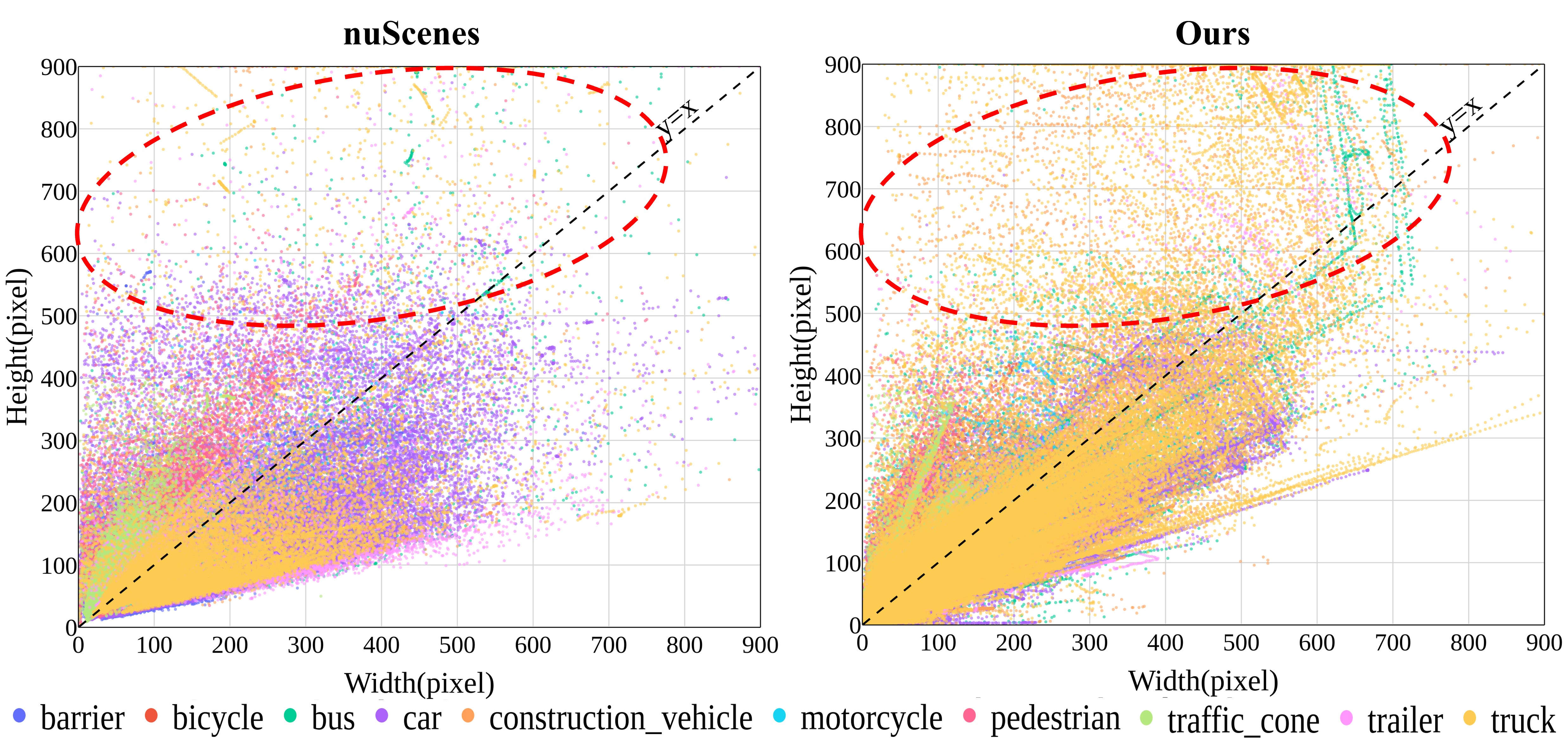}
  \vspace*{-5mm}
  \caption{Scatter plot of 2D-BBox labels for front-camera images in \nuscenesdataset{} (train split) and \moraidataset{}. The colors highlight the object category of individual BBox labels. For visibility, the x-axis (width) of plot is clipped at 900 which is the largest value that height of 2D BBox can take. Lastly, BBoxes with area lower or equal to 10 are excluded from visualization.}
  \label{fig:nusc_morai_scatter_plot}
  \vspace{-10pt}
\end{figure}

Similarly, \reffigure{fig:nusc_morai_scatter_plot} depicts the distribution of 2D BBoxes in image space for each dataset where the individual 2D labels are represented as circular markers at coordinates corresponding to their respective widths and heights. 
In general, the two distributions show a similar tendency and coverage in the width-height 2D space. 
However, it is noteworthy that the distribution of our dataset shows higher density in the area highlighted by the red dotted ellipse where BBox labels with high values for height (500+) and a wide range for width (0 \textasciitilde{} 800) exist. This suggests that images in our dataset contain more diverse shapes of huge vehicles such as buses, construction vehicles, trucks, and trailers, likely at close distances. Combining all the above, mitigation of the class imbalance and enhancement of data diversity could be expected when adding \moraidataset{} to \nuscenesdataset{}.

\begin{table*}[ht]
\caption{Quantitative evaluation of performance gain that \moraidataset{} (\synaux{}) delivers on both \nuscenesdataset{} (\realsource{}) and \textbf{unforeseen} \aihubdataset{} (\realtarget{}). C and L denote Camera and LiDAR, respectively. CV and MC are construction vehicle and motorcycle.}
\label{tab:perception_experiment_overview}
\centering
\begin{threeparttable}[b]
\resizebox{\textwidth}{!}{
    \begin{tabular}{lllll|cccccc|cc|cccc|cc} 
    \hline
    \toprule[1.5pt]
    \multirow{3}{*}{\bfseries{Task}} & \multirow{3}{*}{\bfseries{Modality}} & \multirow{3}{*}{\bfseries{Model}} &\multirow{3}{*}{\bfseries{Remark}} &  \multirow{3}{*}{\bfseries{Training Dataset}} & \multicolumn{14}{c}{\bfseries{Evaluation Dataset}} \\
    \cline{6-19}
    \noalign{\vskip 2pt} 
    & & & & & \multicolumn{8}{c|}{{\bfseries \nuscenesdataset{} (\realsource{})}} & \multicolumn{6}{c}{{\bfseries \aihubdataset{} (\realtarget{})}} \\ 
    \noalign{\vskip 2pt} 
    \cline{6-19}
    & & & & & Truck & Bus & Trailer & CV & MC & Bicycle &\bfseries{\map{}}\tnote{\textdagger{}} & \bfseries{\nds{}} & Car & Truck & Bus & Ped & \bfseries{\map{}}\tnote{\textdaggerdbl{}} & \bfseries{\nds{}} \\
    \bottomrule[1.5pt]
    \noalign{\vskip 2pt} 
    2D & C & \fasterrcnn & \xmark & \nuscenesdataset{}  & 31.4 & 48.0 & 16.1 & 4.2 & 20.9 & 19.8 & 27.3 & \xmark & 19.5 & 13.8 & 16.8 & 3.3 & 13.35 & \xmark  \\
    2D & C & \fasterrcnn & \xmark & \nuscenesdataset{}  + \moraidataset{} & 31.0 & 50.0 & 16.5 & 3.7 & 22.0 & 20.7 & \textbf{27.8} & \xmark & 27.2 & 18.6 & 24.5 & 8.3 & \textbf{19.7} & \xmark \\
    \cline{6-19}
    \noalign{\vskip 1pt} 
    & & & & & \degrade{0.4} & \improve{2.0} & \improve{0.4} & \degrade{0.5} & \improve{1.1} & \improve{0.9} & \improve{0.5} & \xmark & \improve{7.7} & \improve{4.8} & \improve{7.7} & \improve{5.0} & \improve{6.35} & \xmark \\ 
    
    \hline
    \noalign{\vskip 2pt} 
    3D & L & \pointpillars & SECFPN & \nuscenesdataset{}  & 39.2 & 51.7 & 28.6 & 5.4 & 21.5 & 0.9 & 34.94 & 50.02 & 24.93 & 3.425 & 11.70 & 10.93  & 12.74 & 21.95 \\
    3D & L & \pointpillars & SECFPN  & \nuscenesdataset{}  + \moraidataset{} & 40.5 & 53.1 & 30.8 & 7.3 & 22.9 & 0.8 & \textbf{35.45} & \textbf{50.44} & 29.67 & 5.59  & 12.28 & 11.99 & \textbf{14.88} & \textbf{22.63}  \\
    \cline{6-19}
    \noalign{\vskip 1pt} 
    & & & & & \improve{1.3} & \improve{1.4} & \improve{2.2} & \improve{1.9} & \improve{1.4} & \degrade{0.1} & \improve{0.51} & \improve{0.42} & \improve{4.74} & \improve{2.165} & \improve{0.58} & \improve{1.06} & \improve{2.14} & \improve{0.68}  \\ 
    
    \hline
    \noalign{\vskip 2pt} 
    3D & L & \ssn & \xmark & \nuscenesdataset{} & 49.5 & 65.8 & 33.7 & 17.1 & 52.6 & 23.3 & 48.29 & 59.42 & 27.30  & 9.93 & 19.94 & 15.48  & 18.16 & 26.92  \\
    3D & L & \ssn & \xmark & \nuscenesdataset{} + \moraidataset{} & 52.2 & 66.4 & 33.1 & 16.8 & 52.5  & 24.6 & \textbf{48.92} & \textbf{60.41} & 36.76 & 13.79 & 24.16 & 20.76 & \textbf{23.87} & \textbf{30.21}  \\
    \cline{6-19}
    \noalign{\vskip 1pt} 
    & & & & & \improve{2.7} & \improve{0.6} & \degrade{0.6} & \degrade{0.3} & \degrade{0.1} & \improve{1.3} & \improve{0.63} & \improve{0.99} & \improve{9.46} & \improve{3.86} & \improve{4.22} & \improve{5.28} & \improve{5.71} & \improve{3.29}  \\ 

    \hline
    \noalign{\vskip 2pt} 
    3D & L & \centerpoint & Pillar (0.2m) & \nuscenesdataset{} & 49.0 & 63.3 & 31.4 & 10.9 & 41.4 & 18.6 & 48.96 & 59.42 & 0.0{\tnote{*}}& 0.0{\tnote{*}} & 0.065 & 0.0{\tnote{*}} & 0.016 & 2.64  \\
    3D & L & \centerpoint & Pillar (0.2m) & \nuscenesdataset{} + \moraidataset{} & 49.5 & 64.3 & 32.8 & 14.3 & 44.4& 13.9 & \textbf{49.43} & \textbf{59.76} & 14.73 & 0.012 & 2.01 & 0.0{\tnote{*}} & \textbf{4.19} & \textbf{11.38}  \\
    \cline{6-19}
    \noalign{\vskip 1pt} 
    & & & & & \improve{0.5} & \improve{1.0} & \improve{1.4} & \improve{3.4} & \improve{3.0} & \degrade{4.7} & \improve{0.47} & \improve{0.34} & \improve{14.73} & \improve{0.012} & \improve{1.945} & 0.0 & \improve{4.17} & \improve{8.74} \\ 
    
    \hline
    \noalign{\vskip 2pt} 
    3D & L & \centerpoint & Voxel (0.1m) & \nuscenesdataset{} & 53.2 & 66.5 & 36.0 & 15.0 & 55.2 & 36.8 & 56.23 & 64.51 & 0.717 & 0{\tnote{*}} & 0{\tnote{*}} & 0{\tnote{*}} & 0.179 & 7.203  \\
    3D & L & \centerpoint & Voxel (0.1m) & \nuscenesdataset{}  + \moraidataset{} & 55.1 & 69.2 & 37.3 & 18.5 & 57.0 & 38.1 & \textbf{57.80} & \textbf{65.47} & 63.44 & 5.682 & 4.264 & 0{\tnote{*}} & \textbf{18.34} & \textbf{22.93}   \\
    \cline{6-19}
    \noalign{\vskip 1pt} 
    & & & & & \improve{1.9} & \improve{2.7} & \improve{1.3} & \improve{3.5} & \improve{1.8} & \improve{1.3} & \improve{1.57} & 
    \improve{0.96} & \improve{62.72} & \improve{5.682} & \improve{4.264} & 0.0 & \improve{18.16} & \improve{15.72}  \\ 

    \hline
    \noalign{\vskip 2pt} 
    3D & L & \centerpoint & Voxel (0.075m) & \nuscenesdataset{} & 55.1 & 67.9 & 34.9 & 15.2 & 55.8 & 36.1 & 56.95 & 65.40  & 0.197 & 0.0{\tnote{*}} & 0.0{\tnote{*}} & 0.0{\tnote{*}} & 0.049 & 4.19  \\
    3D & L & \centerpoint & Voxel (0.075m) & \nuscenesdataset{} + \moraidataset{}  & 55.9 & 67.1 & 35.7 & 16.1 & 58.4 & 40.1 & \textbf{58.02} & \textbf{66.02} & 63.28 & 1.0 & 24.6 & 0.0{\tnote{*}} & \textbf{22.22} & \textbf{23.71}  \\
    \cline{6-19}
    \noalign{\vskip 1pt} 
    & & & & & \improve{0.8} & \degrade{0.8} & \improve{0.8} & \improve{0.9} & \improve{2.6} & \improve{4.0}& \improve{1.07} & \improve{0.62} & \improve{63.08} & \improve{1.0} & \improve{24.6} & 0.0 & \improve{22.17} & \improve{19.52} \\ 

    \bottomrule[1.5pt]
    \end{tabular}
}
\begin{tablenotes}[para]
\footnotesize
\item [(\textdagger{})] Over 10 detection classes of \nuscenesdataset{}. \item [(\textdaggerdbl{}{})] Over 4 classes: car, truck, bus, and pedestrian. \\ \item [(*)] If the precision or recall values for all operating points on the precision-recall curve are less than 10$\%$, the AP for that class is set to zero \cite{nuscenes}.
\end{tablenotes}
\end{threeparttable}
\vspace{-15pt}
\end{table*}

\section{EXPERIMENTS}

\subsection{Impact of \moraidataset{} on \realsource{} (\nuscenesdataset{} dataset)} \label{exp:source_domain_od}
This section presents quantitative results that \moraidataset{} could maintain or even improve the performance of camera-based 2D and LiDAR-based 3D detectors on \nuscenesdataset{}.

\subsubsection{Pre-processing} \label{source_domain_od_data_preprocessing} We convert \moraidataset{} dataset to follow the format of \nuscenesdataset{} (\eg{} structure, file hierarchy, and keyframe-sweep), primarily meaning annotation frequency of \moraidataset{} decrease from 20 Hz to 2 Hz. 
This step is necessary as numerous modern 3D detectors trained on \nuscenesdataset{}, especially LiDAR-based ones, exploit the sweeps (frames w/o annotations) from past time steps for improving point cloud density and robustness of temporal information. 
Thus, to construct a coherent training environment for those detectors we conduct the format conversion although it decreases the frequency of \moraidataset{} (keyframes: \textasciitilde{}{}3.7K, sweeps: \textasciitilde{}33K).

\subsubsection{Training details} No advanced training or domain-adaptation strategy is applied other than a simple concatenation of \nuscenesdataset{} and \moraidataset{} to construct a merged training set. For monocular 2D detection scenario, \fasterrcnn{} \cite{Ren_2017} is opted. We employ the implementation of \fasterrcnn{} from MMDetection \cite{mmdetection}, and the default training configuration (e.g., learning rate, batch size) remains unchanged. 
The detector is trained on one RTX 3090 GPU with images that are from the front camera. 
For LiDAR-based 3D detection, renowned \pointpillars{} \cite{lang2019pointpillars}, \ssn{} \cite{zhu2020ssn}, and \centerpoint{} \cite{yin2021center} are employed. 
we use the implementation of the mentioned networks in MMDetection3D \cite{mmdet3d2020}. 
While the other configurations are the same as the default ones, batch size is adjusted to our GPU resources.
In short, \pointpillars{}, and \ssn{} are trained on a single RTX 3090 with batch size of 8.
\centerpoint{} is trained on four NVIDIA L4 GPUs with batch sizes set to 8, 8, and 4 for Pillar (0.2m) and Voxel (0.1m, 0.075m), respectively.

\subsubsection{Metrics} We use renowned average precision (AP), mean average precision (\map{}), and nuScenes detection score (\nds{}) \cite{nuscenes}. All metrics are represented in percentage ($\%$), and higher scores indicate better performance. 

\subsubsection{Evaluation} The left half of \reftable{tab:perception_experiment_overview} presents performance of detectors on the validation split of \nuscenesdataset{}. 
As demonstrated, detectors trained on \nuscenesdataset{} combined with \moraidataset{} outperform all baselines that are trained on \nuscenesdataset{} alone in terms of \map{} and \nds{} across modalities and network architectures. While the \map{} gains achieved in the other detectors are marginal, \moraidataset{} brings notable improvement of 1+$\%$ to \centerpoint{}-Voxel (0.1m, 0.075m) models. This gain is worth mentioning as no tailored training strategy was applied to close the sim-to-real gap explicitly.

In addition to \map{}, we present class-wise AP for the rare categories that \moraidataset{} mitigates long-tail issues discussed in \refsection{sec:comp_to_nusc}. Although the trend of AP improvement or decline varies across models for all six classes, we observe meaningful enhancements, \eg{} +2.7$\%$ for truck in \ssn{}, +3.4$\%$ and +3.0$\%$ for CV and MC in \centerpoint{}-Pillar, and +4.0$\%$ for bicycle in \centerpoint-Voxel(0.075m). Combining all the above leads us to conclude the characteristics of \nuscenesdataset{} we benchmark are well implemented in \moraidataset{}, and it does not degrade detectors' performance on \nuscenesdataset{} overall.    

\subsection{Impact of \moraidataset{} on \realtarget{} (\aihubdataset{} dataset)} 
\label{exp:target_domain_od}
To validate the efficacy of simulated features provided by \moraidataset{} in terms of previewing and indirectly learning about \realtarget{}, we load the trained models from \refsection{exp:source_domain_od} and evaluate their performance on \realtarget{} without further training. 

\begin{table}[tb]
\caption{Summary of properties by dataset. Arrow($\rightarrow$) indicates the applied pre-processing for a coherent experiment setup. \\ T and V denote Train and Validation, respectively.}
   \label{tab:morai_nusc_aihub_comparison}
   \centering
\resizebox{0.45\textwidth}{!}{%
  \begin{threeparttable}[b]
   \begin{tabular}{l|cccccc}
    \toprule[1.5pt]
      & \bfseries{$\#$Annot. Frame} & \bfseries{$\#$CAM.} & \bfseries{Resolution}&  \bfseries{$\#$Beam} & \bfseries{$\#$Det. Class} & \bfseries{Freq. (Hz)} \\
     \hline
     \rule{0pt}{3ex}    
      \nuscenesdataset{} (T) & 28K  & 6$\rightarrow$1 &1600$\times$900  & 32 & 10 & 2\\
     \rule{0pt}{4ex}    
    \textbf{\moraidataset{} (T) } & 37K$\rightarrow$3.7K  & 6 $\rightarrow$1 &1600$\times$900 & 32 & 10 &20 $\rightarrow$2\\
     \hline
     \rule{0pt}{4ex}    
    \aihubdataset{} (V) & 80K$\rightarrow$8.8K & 5$\rightarrow1$ & 1920$\times$1200 $\rightarrow$1600$\times$900 & 128 & 8$\rightarrow$4 & 10$\rightarrow$2\\
    \bottomrule[1.5pt]
     \end{tabular}
  \end{threeparttable}
}
\vspace{-15pt}
\end{table}

\subsubsection{Real-target dataset} A real-world driving dataset, \href{https://aihub.or.kr/aihubdata/data/view.do?dataSetSn=71590}{\aihubdataset{}} is opted to represent \realtarget{} given (1) its sensor data are collected from the real-target domain of this paper, South Korea where digital twin maps in \moraidataset{} (\synaux{}) are originated from. (2) Sensor suite and GT types are similar to \nuscenesdataset{} and \moraidataset{}, \ie{} each frame contains five 1920$\times$1200 multi-view images, one point cloud from 128-channel spinning LiDAR, and RTK GNSS data with 3D-BBox annotation frequency of 10 Hz.

\subsubsection{Pre-processing}
\reftable{tab:morai_nusc_aihub_comparison} summarizes the applied modifications.
Format of \aihubdataset{} dataset is converted to \nuscenesdataset{} style, following \refsection{source_domain_od_data_preprocessing}. On top of that, we apply additional processing for fair experiments with \refsection{exp:source_domain_od}. (1) Only images from the front camera are used for 2D detection, and they are cropped to 1600$\times$900. (2) We generated pseudo-2D-BBox labels exploiting pixel-level panoptic segmentation GT as they are not present in \aihubdataset{}. Lastly, (3) we take 3D BBoxes only for four classes out of eight, \ie{} car, truck, bus, and pedestrian, as they either have deviating labeling policies from \nuscenesdataset{} or have few instances ($\leq$ 100). Note that all data of \aihubdataset{} are used for validation, not training.

\subsubsection{Evaluation} The right half of \reftable{tab:perception_experiment_overview} shows detection scores of models evaluated on the unobserved \aihubdataset{} dataset. 
As can be seen, detectors trained with \moraidataset{} exceed respective baselines by a large margin across modalities. 
For example, \map{} gains of 6.3$\%$ for camera-based \fasterrcnn{} and 5.7$\%$ for LiDAR-based \ssn{} are reported. 
This suggests that \moraidataset{} can provide simulated real-world experiences, assisting the adaptation of detectors and lessening reliance on real sensor data from \realtarget{}.

\begin{figure}[tb]
  \centering
    \includegraphics[width=0.45\textwidth]{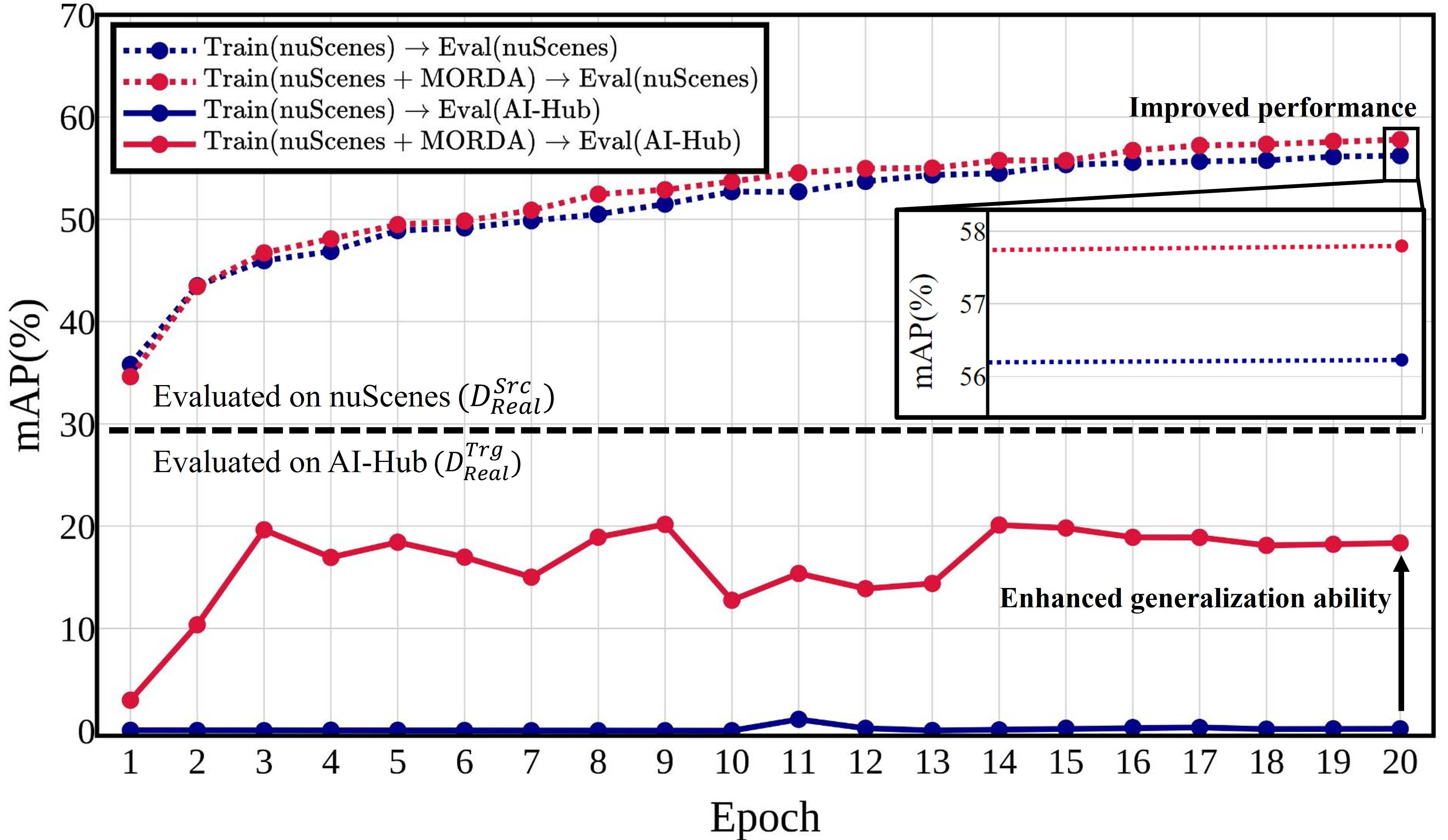}
  \vspace*{-4mm}
  \caption{Development of the \map{} scores, training \centerpoint{}-Voxel (0.1m) over 20 epochs w/o (\blue{blue}) and w/ (\red{red}) \moraidataset{}. The same models are evaluated on \nuscenesdataset{} (dotted) and \aihubdataset{} (solid), respectively.} 
  \label{fig:id_ood_over_epochs}
  \vspace{-15pt}
\end{figure}

\begin{figure}[tb]
  \centering
    \includegraphics[width=0.5\textwidth]{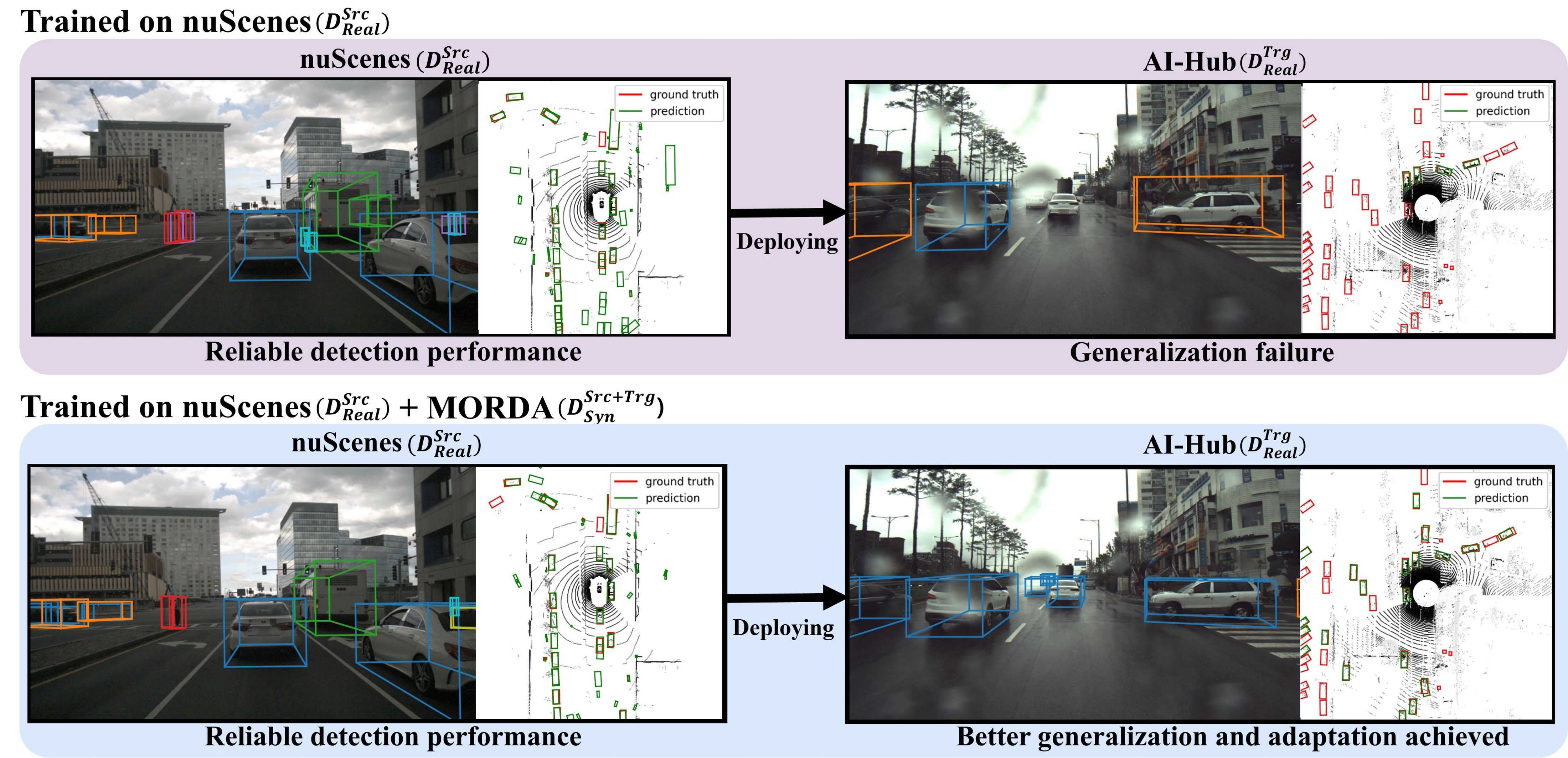}
  \vspace*{-6mm}
  \caption{Enhanced generalization performance of \centerpoint{}-Voxel (0.1m) with aid of \moraidataset{}. Camera images overlay predictions from the models.} 
  \label{fig:enhanced_adaptation_to_aihub}
  \vspace{-5pt}
\end{figure}

In addition to the mentioned benefit above, we observe \moraidataset{} can assist in mitigating the generalization failure. 
\reftable{tab:perception_experiment_overview} shows that the baseline \centerpoint{} family reports severe performance degradation under domain shift (\nuscenesdataset{}$\rightarrow$\aihubdataset{}), meaning their class-wise AP and \map{} scores on \aihubdataset{} almost reach to zero regardless of backbone types. 
In contrast, when \moraidataset{} is included, \map{} scores of all \centerpoint{} models get improved, meaning the observed generalization failures get alleviated. Especially, those with voxel backbones (0.1m and 0.075m) gain significant improvements in \map{} (+18$\%$ and +22$\%$).
\reffigure{fig:id_ood_over_epochs} explores this further and illustrates the development of \map{} throughout entire training process with respect to the presence of \moraidataset{}. Notice that the model with \moraidataset{} achieves a substantial performance gap compared to the baseline on \aihubdataset{}. In light of the analysis above, we conclude that \moraidataset{} is a well-constructed preview for \realtarget{} and a useful regularization tool for generalization failure.
\reffigure{fig:enhanced_adaptation_to_aihub} illustrates benefits of \moraidataset{} in this aspect.

\subsection{Benchmark with Other Synthetic Datasets: 2D Detection}  \label{exp:benchmark_2d_od}
\virtualkittiseconddataset{}, \synthiadataset{}, and \shiftdataset{} datasets are compared to \moraidataset{} in terms of mAP gains that respective datasets bring to \fasterrcnn{} trained on \nuscenesdataset{}.
\subsubsection{Pre-processing} We use images from the front camera that all mentioned datasets above have in common. Next, we sample the datasets to match their frequencies as much as possible. That means \moraidataset{}, \synthiadataset{}, and \virtualkittiseconddataset{}\footnote{The frequency of \virtualkittiseconddataset{} is assumed to be the same as \kittidataset{}, as we could not find relevant information.} are downsampled to 2Hz. Although images of \shiftdataset{} dataset (discrete-shift) are already of 1 Hz, we downsample it by a factor of 5 due to its exceptional size (150K). Note that \shiftdataset{} is still the largest dataset after this downsampling, as shown in \reftable{tab:map_across_datasets}. 2D BBoxes of respective datasets are merged to \nuscenesdataset{} based on class labels.
\subsubsection{Training Details} \fasterrcnn{} is trained on \nuscenesdataset{} and either of the synthetic datasets for all 10 detection classes, using MMDetection with default configuration. 
\subsubsection{Evaluation} 
\reftable{tab:map_across_datasets} presents \map{}s of \fasterrcnn{}s trained with respective synthetic datasets along with the numbers of frames used for training. Our \moraidataset{} achieves 27.8$\%$, surpassing the baseline and all the compared datasets. \shiftdataset{} shows a comparable mAP gain with ours, but it used 8$\times{}$ more frames than \moraidataset{}. 

\begin{table}[tb]
\centering
\caption{\map{} gains in \fasterrcnn{} by synthetic dataset. Detection performance is assessed on validation split of \nuscenesdataset{}.}
\resizebox{0.37\textwidth}{!}{%
  \begin{threeparttable}[b]
   \label{tab:map_across_datasets}
   \begin{tabular}{ll|c}
    \toprule[1.5pt]
     \bfseries{Training Dataset} & \bfseries{$\#$Frame} & \bfseries{\map{} ($\%$, $\uparrow$)} \\
     \bottomrule[1.5pt]
     \noalign{\vskip 2pt} 
     \nuscenesdataset{} & 28K & 27.3 \\
     \hline
     \nuscenesdataset{} + \virtualkittiseconddataset{} & 28K + 2.5K & 27.0 (\degrade{0.3})\\
     \nuscenesdataset{} + \synthiadataset{} & 28K + 7.3K & 27.4 (\improve{0.1}) \\ 
     \nuscenesdataset{} + \shiftdataset{} & 28K + 30K & 27.7 (\improve{0.4})\\ 
     \textbf{\nuscenesdataset{} + \moraidataset{} (Ours)} & 28K + 3.7K& \textbf{27.8 (\improve{0.5})}\\ 
    \bottomrule[1.5pt]
     \end{tabular}
  \end{threeparttable}
}
\vspace{-15pt}
\end{table}

\section{CONCLUSIONS}
For safety-critical AVs, DNNs need to overcome the performance degradation that occurs in unexperienced \realtarget{} while maintaining performance on \realsource{}. 
This paper proposed leveraging \synaux{} where key characteristics of \realsource{} and \realtarget{} are fused. To showcase the efficacy, we generated a novel synthetic dataset, \moraidataset{}. 
Our comprehensive experiments demonstrate that \moraidataset{} could help DNNs adapt to \realtarget{}, mitigating their dependency on real-world data without compromising performance on \realsource{}. 

Nevertheless, there is room for further research. Since only a simple concatenation of \nuscenesdataset{} and \moraidataset{} was used in this study, further performance enhancement on \realtarget{} might be achieved by additionally incorporating modern UDA methods. Another direction for future work could be to identify which characteristics implemented in \moraidataset{} contributed to performance stability on \realtarget{} by conducting an in-depth comparison with other synthetic datasets.

\bibliographystyle{IEEEtran}
\bibliography{IEEEabrv,mybibfile}

\end{document}